\pdfoutput=1
\documentclass[letterpaper]{article}
\usepackage[preprint]{aaai2027}
\usepackage[hyphens]{url}  
\usepackage{graphicx} 
\usepackage{natbib}  
\usepackage{caption} 
\usepackage{algorithm}
\usepackage{algorithmic}

\usepackage{booktabs}

\usepackage{amsmath,amssymb}
\usepackage{etoolbox}

\makeatletter
\patchcmd{\@maketitle}{These authors contributed equally.}{Equal contribution.}{}{}
\patchcmd{\@maketitle}{These authors contributed equally.}{Equal contribution.}{}{}
\makeatother

\newcommand{\PublicAuthors}{
Yi Wei\textsuperscript{\rm 1,\rm 2}\equalcontrib,
Shuo Jiang\textsuperscript{\rm 1}\equalcontrib,
Huaixia Dou\textsuperscript{\rm 1},
Jie Zhu\textsuperscript{\rm 1}\corresponding,
Junhui Li\textsuperscript{\rm 3}, \\
Lifan Guo\textsuperscript{\rm 1},
Feng Chen\textsuperscript{\rm 1},
Chi Zhang\textsuperscript{\rm 1}
}
\newcommand{\PublicAffiliations}{
\textsuperscript{\rm 1}Qwen DianJin Team, Alibaba Cloud Computing\\
\textsuperscript{\rm 2}Beihang University\\
\textsuperscript{\rm 3}School of Computer Science and Technology, Soochow University
}
\newcommand{\PublicPdfAuthors}{Yi Wei, Shuo Jiang, Huaixia Dou, Jie Zhu, Junhui Li, Lifan Guo, Feng Chen, Chi Zhang}

\title{Dual-Loop Self-Evolution via Verifiable Emotion Feedback \\ for Multi-Turn Empathetic Dialogue}
\author{\PublicAuthors}
\affiliations{\PublicAffiliations}

\begin{document}

\maketitle

\begin{abstract}
Large language models have demonstrated conversational capabilities, yet empathetic competence remains challenging. Empathetic support is inherently multi-turn and path-dependent: users disclose concerns gradually, emotions evolve over time, and early responses shape trust and receptivity. Reinforcement learning with verifiable emotion rewards provides scalable supervision for long-horizon interactions. However, existing methods evolve the dialogue policy while keeping its training interaction distribution fixed, creating a mismatch between policy competence and training experience. We introduce a dual-loop self-evolution framework driven by verifiable emotion feedback. With the user simulator and verifier frozen, the inner loop optimizes the multi-turn policy using continuous emotion rewards, while the outer loop uses the same outcomes to estimate policy-relative interaction utility and adapt experience. To obtain estimates from sparse, stochastic rollouts, the framework holds the scenario and interaction state constant within each group and prioritizes conditions whose group pass rates lie near the policy's competence boundary. A hierarchical controller shares evidence across support intents, while uncertainty-guided exploration and uniform rehearsal prevent premature exclusion. The resulting distribution generates trajectories, closing both loops without increasing the rollout budget. On SAGE, our framework raises Qwen3-8B Overall from 53.87 to 79.24 and outperforms protocol-matched uniform emotion-reward reinforcement learning by 7.23 points.
\end{abstract}

\section{Introduction}
Large language models (LLMs) have become capable conversational systems, yet reliable empathetic support remains difficult \cite{rashkin2019empathetic,esconv2021,multiesc2022}. Empathy is inherently multi-turn: concerns surface incrementally over turns, affect rises and falls with the conversation, and each response constrains what the user is willing to reveal next \cite{supporter2023,emodynamix2025,wutraj2026,sage2025}. A reply that appears appropriate in isolation may still derail the subsequent conversation. An effective policy must therefore learn how its actions influence an evolving user over a complete trajectory, rather than merely imitate locally plausible supportive text.

\begin{figure}[!t]
\centering
\includegraphics[width=\columnwidth]{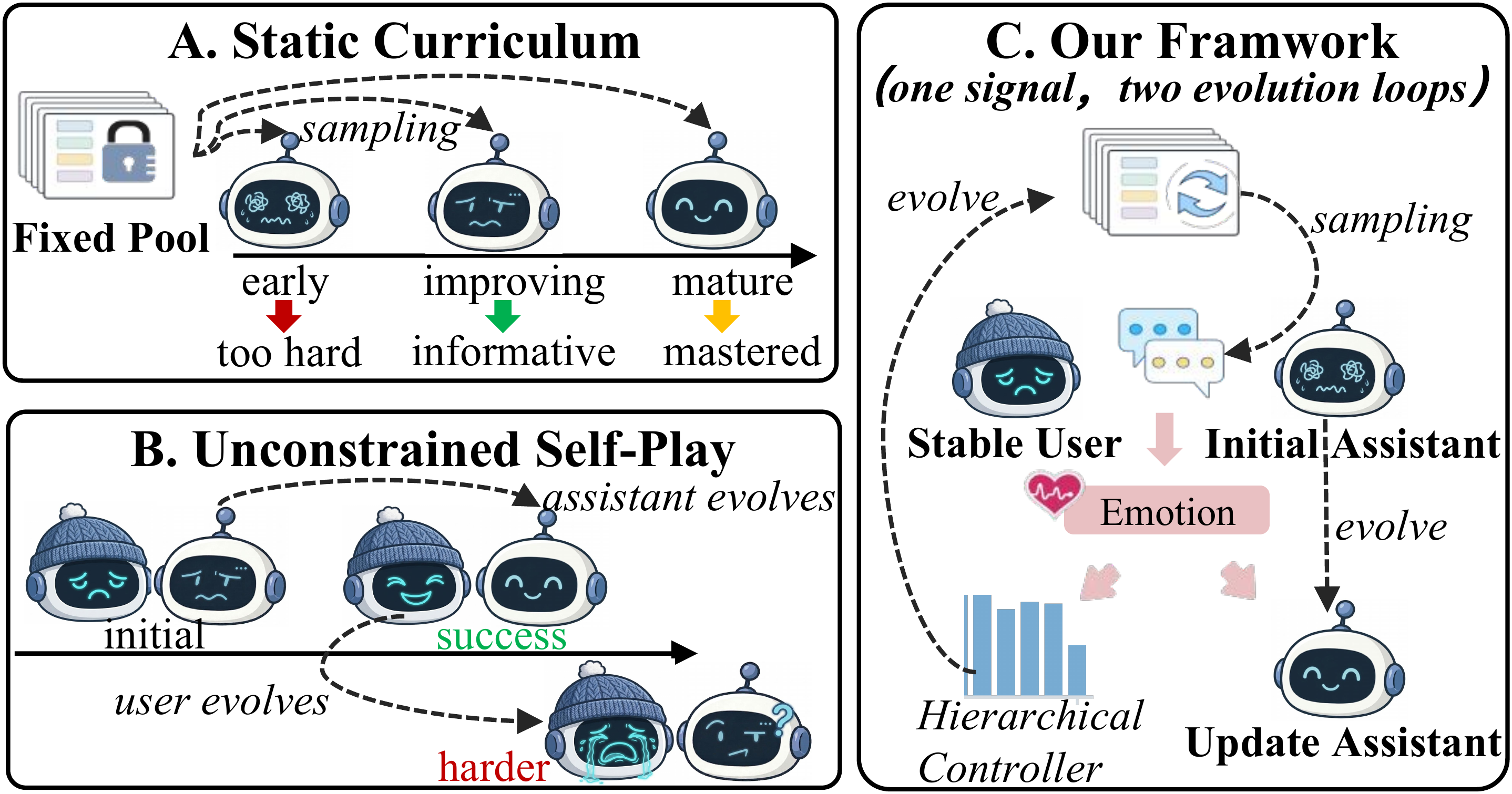}
\caption{Training paradigms. Static curricula lag behind competence; unconstrained self-play creates moving targets; ours adapts the interaction distribution with stable role-play.}
\label{fig:motivation}
\end{figure}

Learning such behavior requires coherent long-horizon experience and reliable outcome supervision. Human-authored support dialogues are costly to collect, difficult to standardize, and limited in the range of user reactions they expose \cite{esconv2021,multiesc2022}. Offline augmentation improves coverage but remains fixed once generated \cite{augesc2023,ye2025generic}. Interactive user simulation offers a scalable alternative: a policy can observe how different responses change later disclosure and emotion, provided the simulated user preserves its persona, context, and hidden need across turns \cite{yoon2024usersim,gromada2025persona,dou2025simulatorarena,sage2025}. Verifiable emotion rewards further turn these reactions into trajectory-level supervision, enabling reinforcement learning (RL) to optimize long-term affective consequences rather than only match static references \cite{rlver2025}.

However, existing methods still treat the training interaction distribution as a predefined and fixed external condition \cite{rlver2025}. Emotion rewards continually update the dialogue policy, but do not change how subsequent experience is allocated. Consequently, each condition receives the same expected rollout budget whether the current policy has mastered it, repeatedly fails on it, or still finds it informative. As competence evolves, the learning value of an interaction changes with it, while the allocation of expensive multi-turn experience remains static. This creates a growing mismatch between what the policy can currently do and what it is asked to practice.

Simply enlarging the interaction pool or prescribing an easy-to-hard schedule does not resolve this mismatch. Interaction difficulty is not an intrinsic, permanent property of a scenario. The same hesitant or emotionally activated user may be excessive for an early policy, informative for an improving policy, and redundant for a mature one. A curriculum fixed before training cannot remain aligned with this moving competence boundary \cite{bengio2009curriculum,kumar2010selfpaced}. Adaptive curriculum and replay methods recognize that training utility is learner-dependent \cite{graves2017acl,schaul2016per,jiang2021plr,adarft2025,vcrl2025}, but existing formulations do not directly address sparse, stochastic outcomes from long multi-turn interactions with a role-playing user.

Self-play offers an apparently natural solution by evolving both the assistant and its simulated partner \cite{chen2024spin,sead2026}. In empathetic dialogue, however, the simulator participates in both trajectory generation and outcome formation. Unconstrained adaptation can make users arbitrarily resistant, overly accepting, or inconsistent with their original persona and hidden need \cite{yoon2024usersim,dou2025simulatorarena}. Reward changes would then reflect both assistant improvement and simulator drift, obscuring credit assignment and turning evaluation into a moving target. A more difficult user is not necessarily a more useful or faithful teacher.

To this end, we propose a dual-loop self-evolution framework that jointly advances the dialogue policy and the experience distribution through shared verifiable emotion feedback; Figure~\ref{fig:motivation} contrasts this design with static curricula and unconstrained self-play. The loops are nested but have distinct targets: the inner loop uses continuous emotion rewards to improve \emph{how the policy responds}, while the outer loop reuses group outcomes to identify \emph{what the policy should practice next} and reallocates subsequent interactions accordingly. The updated distribution produces new trajectories that again supervise both processes, turning a fixed interaction pool into a capability-aligned online curriculum grounded in comparable multi-turn outcomes.

Experiments show that our framework achieves 79.24 SAGE Overall, surpassing uniform emotion-reward RL by 7.23 points, or a 10.0\% relative gain. The improvement requires no additional rollouts and holds across complementary evaluation protocols and component ablations.

Our contributions are threefold:
\begin{itemize}
  \item We identify \emph{curriculum--policy misalignment} as a central challenge in multi-turn emotion-reward RL: policy competence evolves while its interaction distribution remains fixed.
  \item We introduce a dual-loop self-evolution framework in which shared emotion feedback jointly drives policy improvement and capability-aligned adaptation of the training experience distribution.
  \item Multi-run, cross-benchmark, and component-level evaluations demonstrate substantial performance gains and improved rollout-budget efficiency over uniform emotion-reward training.
\end{itemize}

\section{Related Work}
\paragraph{Empathetic dialogue and affective reinforcement learning.}
Empathetic dialogue has progressed from emotion-grounded generation to long-horizon support over latent needs and strategies \cite{rashkin2019empathetic,lin2019moel,majumder2020mime,esconv2021,multiesc2022}, with work on planning, augmentation, interpretable reasoning, strategy prediction, and personalization \cite{supporter2023,augesc2023,escot2024,kang2024supporter,emodynamix2025,shi2025beyond,ye2025generic}. RL encourages value-sensitive support \cite{kim2025value,cso2025,wang2025flexible}, while LLM simulators enable scalable interaction but require faithful role-play and assessment \cite{yoon2024usersim,gromada2025persona,dou2025simulatorarena}. SAGE supplies evolving emotional personas \cite{sage2025}, and verifiable emotion rewards optimize policies \cite{rlver2025}; we instead align the interaction distribution with the evolving policy.

\paragraph{Adaptive curricula and self-evolving training.}
Curriculum, self-paced, and replay methods allocate computation by learner-dependent utility \cite{bengio2009curriculum,kumar2010selfpaced,jiang2015spcl,schaul2016per,jiang2021plr}; LLM self-improvement adds self-play, self-judgment, and online optimization \cite{chen2024spin,yuan2024selfrewarding,ahmadian2024rloo}. Recent work adapts prompts or rollouts using success, uncertainty, variance, and progress \cite{adarft2025,vcrl2025,ducl2026,mracl2026,adacurl2026,cures2026,vip2026}, or evolves tasks, users, evaluators, and harnesses \cite{sead2026,codea12026,coevolve2026,evotrainer2026}. We instead freeze role-play and evolve a hierarchical distribution over multi-turn conditions.

\section{Dual-Loop Self-Evolution}
\begin{figure*}[t]
\centering
\includegraphics[width=\textwidth]{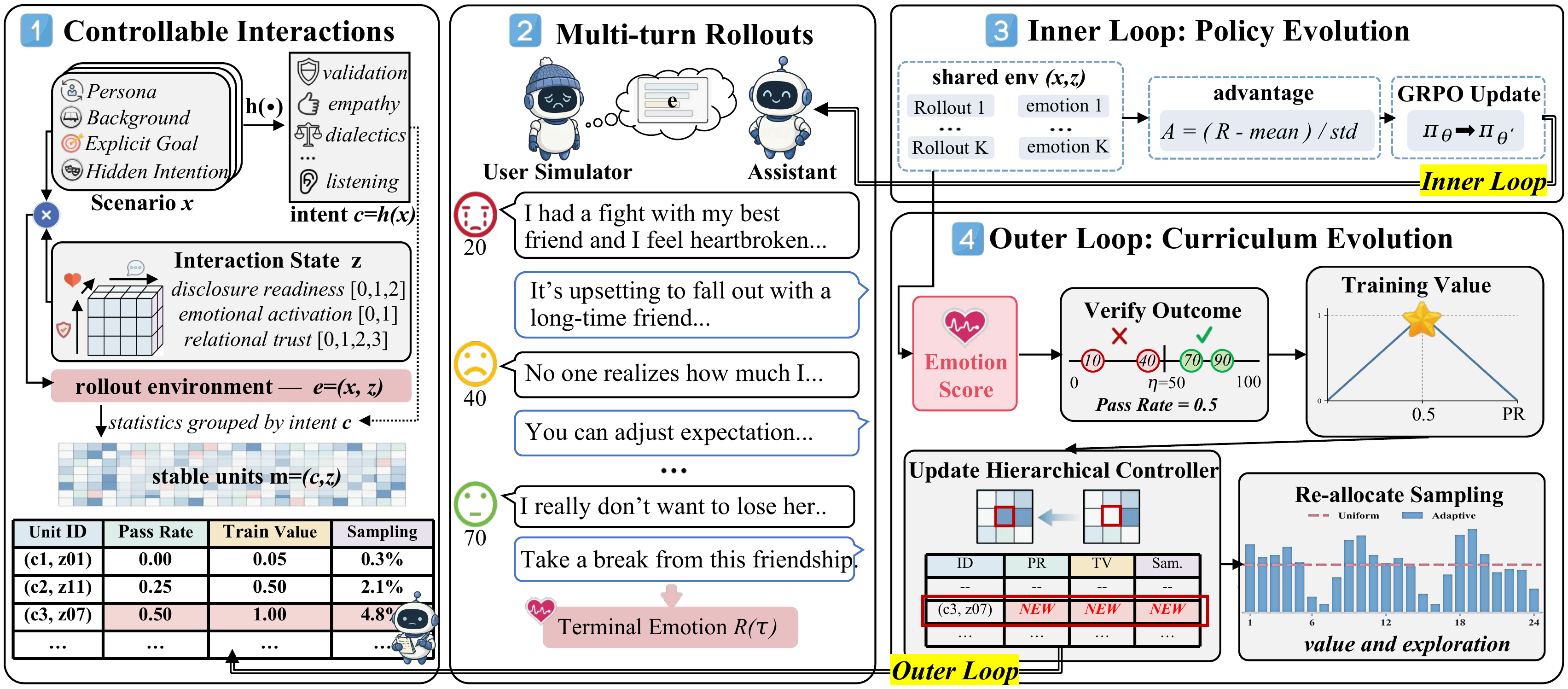}
\caption{Overview of the proposed framework. A complete scenario remains uniformly sampled and is paired with a simulator-side interaction state conditioned on its native support intent. Trajectories within a rollout group share the same environment. Continuous emotion reward optimizes the dialogue policy, while thresholded group outcomes update hierarchical intent--state statistics and the subsequent interaction distribution.}
\label{fig:overview}
\end{figure*}

Figure~\ref{fig:overview} presents the complete training pipeline. Our framework organizes multi-turn empathetic-dialogue training as two \emph{nested} feedback loops driven by the same verified emotion outcomes. The \emph{inner loop} optimizes the policy within the interaction distribution currently supplied by the controller; the \emph{outer loop} observes completed group outcomes and revises that distribution as the policy changes. The two loops are logically nested rather than parallel learners.

The remainder of this section follows one complete feedback cycle. Section~3.1 formalizes scenario-conditioned rollouts and the nested objectives, while Section~3.2 constructs the controllable interaction space. Sections~3.3--3.4 describe shared-condition policy learning and verified group feedback. Section~3.5 converts sparse outcomes into robust policy-relative utility, and Section~3.6 closes the loop by sampling the next interaction distribution.

\subsection{Task Setup and Dual-Loop Framework}
Training begins from a set $\mathcal X$ of complete emotional-support scenarios. A scenario $x\in\mathcal X$ specifies the user's persona, precipitating event, current dilemma, response tendencies, and hidden support need; an intent map $h$ associates it with a support intent $c=h(x)$. We preserve this semantic core throughout training. To control how the same concern unfolds in dialogue, we additionally choose a simulator-side interaction state $z$ from a finite space $\mathcal Z$. The pair $e=(x,z)$ is one concrete rollout environment: $x$ determines \emph{what} the user is experiencing, while $z$ determines \emph{how} the user discloses and reacts.

For a sampled condition $(x,z)$, a user simulator $\mathcal U$ and the dialogue policy $\pi_\theta$ interact for at most $H$ turns. Their alternating user and assistant utterances form a trajectory
\[
\tau=(u_1,a_1,\ldots,u_T,a_T)
\sim P(\tau\mid x,z,\pi_\theta,\mathcal U),
\]
where $T\le H$ is the realized number of turns. After the interaction ends, an emotion verifier evaluates the complete trajectory and returns a scalar outcome $R(\tau)$. The policy observes only the natural-language conversation: support-intent labels, interaction-state labels, controller statistics, and sampling probabilities remain simulator-side metadata. Consequently, improved performance must be expressed through dialogue behavior rather than direct access to a difficulty label.

At training stage $t$, the outer loop represents its current curriculum by a conditional distribution $p_t(z\mid c)$ over interaction states for each support intent, initialized uniformly before feedback accumulates. Sampling from this distribution produces the conditions under which the inner loop optimizes the policy:
\begin{equation}
\begin{aligned}
\max_\theta\quad
&\mathbb E_{x\sim\mathrm{Unif}(\mathcal X),\,
z\sim p_t(\cdot\mid h(x)),\,\tau}\\
&\left[R(\tau)\right].
\end{aligned}
\label{eq:inner}
\end{equation}
Meanwhile, verified outcomes update controller history $\mathcal H_t$ and hence the next interaction distribution:
\begin{equation}
\begin{aligned}
\mathcal H_{t+1}
  &=\operatorname{Update}(\mathcal H_t;\mathcal D_t,R_t),\\
p_{t+1}
  &=\operatorname{Controller}(\mathcal H_{t+1}).
\end{aligned}
\end{equation}
Here $\mathcal D_t$ denotes the completed rollout groups collected at stage $t$, $R_t$ their verified outcomes, and $\mathcal H_t$ the controller's accumulated evidence. Updating $\mathcal H_t$ changes $p_{t+1}$, which in turn changes the experience used by the next policy updates. Crucially, the outer-loop target is policy-relative interaction utility---the value of allocating another rollout group to a condition under the current policy---rather than a permanent difficulty label attached to a scenario.

\subsection{A Controllable Interaction Space}
The framework requires interaction states to be behaviorally realizable, compatible with the underlying scenario, hidden from the assistant, and reusable within a rollout group. We preserve each complete scenario and define three task-operational axes: \emph{disclosure readiness} controls when details and deeper needs are revealed; \emph{emotional activation} controls the intensity of negative-affect expression; and \emph{relational trust} controls whether questions and support are accepted. These axes capture information release, affective expression, and relational feedback rather than clinical traits.

We discretize the axes at a granularity that remains behaviorally distinguishable and statistically estimable: disclosure has three levels (delayed, conditional, proactive), activation has two bounded levels (moderate, high), and trust has four levels from skepticism to sustained engagement. Their Cartesian product gives $|\mathcal Z|=3\times2\times4=24$ interaction states. A state $z=(d,a,r)$ is rendered by composing one natural-language behavioral clause for each axis,
$\phi(z)=\phi_d(d)\oplus\phi_a(a)\oplus\phi_r(r)$ for $z=(d,a,r)$.
These clauses govern disclosure pace, affective expression, and receptivity, but cannot overwrite the persona, event, or hidden need specified by $x$.

Estimating a separate curriculum value for every scenario--state pair would fragment evidence across many nearly related conditions. We therefore retain individual scenarios for rollout generation but organize controller evidence by support intent. Combining intent $c$ with state $z$ defines the reusable unit $m=(c,z)$. This factorization lets different scenarios with the same support objective contribute evidence about a shared interaction state, while preserving intent-specific differences. Scenario identities and intent frequencies remain fixed; only the conditional state distribution $p_t(z\mid c)$ evolves. Section~\ref{sec:ablation} later separates the contribution of this interaction space from that of adaptive allocation.

\paragraph{Behavioral realization and information boundaries.}
Each state is translated into simulator-side behavioral guidance rather than exposed as a symbolic label. Delayed, conditional, and proactive disclosure regulate when relevant facts and deeper needs become available. Moderate and high activation regulate the intensity of negative affect without inventing an unsupported crisis. The four trust levels range from rejecting generic reassurance to sustained engagement with situation-specific support. Every composed instruction additionally requires the simulator to preserve the original persona, event, and hidden support need, and forbids mentioning axis names, levels, state identifiers, or difficulty labels.

\subsection{Group-Shared Multi-Turn Policy Learning}
For each scenario, the controller selects one state $z$, and group-relative policy optimization draws $\tau_{1:K}\overset{\mathrm{i.i.d.}}{\sim}P(\tau\mid x,z,\pi_\theta,\mathcal U)$ under the shared condition $(x,z)$. Holding the environment fixed within a group prevents differences in user behavior from confounding group-relative advantages, so rollout variation primarily reflects policy and generation stochasticity.

Abstracting from stabilization terms held fixed across all controlled systems, the method-level policy reward is the continuous verifier outcome, $r_k^{\mathrm{policy}}=R(\tau_k)$. GRPO computes group-relative advantages from these rewards and applies the standard clipped policy objective with KL regularization \cite{deepseekmath2024,ppo2017}. Building on this inner optimization, our outer allocation loop reuses the same outcomes to organize subsequent training experience.

\subsection{Emotion-Thresholded Group Feedback}
The policy and controller consume different transformations of the same verified outcome. Given an emotion criterion $\eta$, the controller computes
\begin{equation}
y_k=\mathbb I[R(\tau_k)\geq\eta],
\qquad
p_m=K^{-1}\sum_{k=1}^{K}y_k ,
\label{eq:passrate}
\end{equation}
where $y_k$ records whether trajectory $k$ passes the criterion, and $p_m$ is the pass rate of the shared unit $m=(h(x),z)$ in that group. We use the pass rate because it directly reveals whether the current policy produces both successful and unsuccessful behavior under a controlled condition. Thresholding affects only outer-loop allocation: the continuous value $R(\tau_k)$ remains the inner-loop reward and therefore retains fine-grained differences between trajectories. One complete group contributes one pass-rate observation, preventing correlated trajectories from masquerading as independently sampled environments.

\subsection{Robust Policy-Relative Utility Estimation}
Controller evidence is uneven: some intent--state units are observed repeatedly, whereas others have only a few completed groups. Trusting a sparse empirical rate can turn an early random success or failure into a persistent sampling bias. For unit $(c,z)$, let $S_{c,z}$ be the cumulative sum of group pass rates and $N_{c,z}$ the number of complete groups. We stabilize its estimate with a leave-one-intent prior formed from the same state in the other support intents:
\begin{equation}
\begin{aligned}
S_z^{-c}&=\sum_{c'\neq c}S_{c',z},\qquad
N_z^{-c}=\sum_{c'\neq c}N_{c',z},\\
\hat p_z^{-c}&=\frac{S_z^{-c}+\alpha/2}{N_z^{-c}+\alpha},
\qquad
\hat p_{c,z}=\frac{S_{c,z}+\lambda\hat p_z^{-c}}{N_{c,z}+\lambda}.
\end{aligned}
\label{eq:shrink}
\end{equation}
Here $\hat p_z^{-c}$ summarizes how state $z$ behaves outside intent $c$, and $\hat p_{c,z}$ is the resulting unit estimate. The coefficient $\alpha$ supplies a neutral prior when pooled evidence is scarce, while $\lambda$ controls how strongly the unit borrows that evidence. Excluding $c$ avoids counting its observations in both terms. As $N_{c,z}$ grows, the unit's own history naturally dominates.

The stabilized pass rate is then converted into an allocation score. We assign maximal boundary value when success and failure coexist, and add an uncertainty bonus that revisits under-observed units \cite{auer2002ucb}:

\begin{equation}
\begin{aligned}
B_{c,z}&=\max(0,1-2|\hat p_{c,z}-\tfrac12|),\\
U_{c,z}&=\sqrt{\log(N+1)/(N_{c,z}+1)},\\
A_{c,z}&=\max(A_{\min},B_{c,z}+\beta U_{c,z}).
\end{aligned}
\label{eq:boundary}
\end{equation}
Here $B_{c,z}$ measures proximity to the current success boundary, $U_{c,z}$ expresses uncertainty from limited visits, and $A_{c,z}$ combines the two with exploration weight $\beta$ while retaining a minimum score $A_{\min}$. In $U_{c,z}$, $N=\sum_{c',z'}N_{c',z'}$ denotes the total number of completed groups across all units, so that under-observed units receive a larger exploration bonus. Near $\hat p_{c,z}=0.5$, successful and unsuccessful trajectories coexist, providing contrasting behavior under the same interaction condition. Unlike raw reward variance, which is scale-sensitive and can mix learnability with simulator noise or outliers, $B_{c,z}$ is bounded and aligned with the verified success criterion. The controller can therefore estimate policy-relative interaction utility directly from outcomes already produced by training.

The controller first samples states uniformly to establish evidence, then activates feedback-guided allocation. Incomplete groups and groups with non-finite outcomes do not update its history. Thereafter,
\begin{equation}
p_t(z\mid c)=(1-\epsilon)
\frac{A_{c,z}^{1/\gamma}}{\sum_{z'\in\mathcal Z}A_{c,z'}^{1/\gamma}}
+\epsilon\frac{1}{|\mathcal Z|},
\label{eq:sampling}
\end{equation}
where $\gamma$ is a sampling temperature and $\epsilon$ reserves uniform rehearsal. Thus, no state is permanently removed, and conditions that were previously misestimated or excessive can re-enter training as the policy develops.

\subsection{Coupled Self-Evolution}
Every complete group thus closes both feedback paths: continuous outcomes update the policy, thresholded group outcomes update the allocation, and the revised distribution generates the next round of experience. Because the simulator, scenario semantics, and verifier stay frozen, experience evolves without a drifting role-playing target. Algorithm~\ref{alg:egsec} summarizes the end-to-end process.

\begin{algorithm}[t]
\caption{Dual-Loop Self-Evolution with Verifiable Emotion Feedback}
\label{alg:egsec}
\begin{algorithmic}[1]
\REQUIRE scenarios $\mathcal X$ with intent map $h$, interaction states $\mathcal Z$, policy $\pi_\theta$, group size $K$
\FOR{each training batch}
  \FOR{each uniformly sampled scenario $x$}
    \STATE set $c=h(x)$ and sample $z$ uniformly during initialization, otherwise from $p_t(z\mid c)$
    \STATE generate $K$ trajectories under the shared environment $(x,z)$
    \STATE score each trajectory with continuous emotion outcome $R_k$
    \STATE update $\pi_\theta$ using group-relative advantages from $R_{1:K}$
    \STATE set $y_k=\mathbb I[R_k\geq\eta]$ and $p_m=K^{-1}\sum_k y_k$
    \STATE if the group is complete and finite, accumulate $p_m$ for state $z$ and unit $(c,z)$
  \ENDFOR
\ENDFOR
\end{algorithmic}
\end{algorithm}

\section{Experiments}
\subsection{Experimental Setup}
Our experiments answer two questions: whether capability-aligned allocation improves empathetic dialogue under a fixed rollout budget, and whether this improvement comes from adaptive allocation rather than a larger state space or generic sample prioritization. The 500 training scenarios, development split, and 100-scenario SAGE test set are instance-disjoint. Each training scenario retains one of eight native support-intent labels and is composed online with one of 24 interaction states, yielding 192 controller units. Development data is used for configuration and checkpoint selection; test instances never enter policy training, controller updates, or model selection.

We use complementary automatic, interactive, and human evaluations. \textbf{SAGE} measures final user emotion and success/failure over 100 multi-turn scenarios \cite{sage2025}. \textbf{ESC-Eval} evaluates 331 fixed-role interactions with the official InternLM2 and ESC-RANK protocol \cite{zhao2024esceval}. \textbf{EIBench} covers 213 held-out scenarios spanning Support, Defense, Repair, and Charm \cite{eibench2026}. \textbf{ESConv} evaluates 2,895 reference responses using strategy accuracy, BLEU-2, ROUGE-L, BERTScore, and Distinct-2 (all multiplied by 100). To separate training feedback from evaluation, DeepSeek-V3 is used only for training simulation and verification; transfer is measured independently by InternLM2+ESC-RANK in ESC-Eval, Qwen3-Max in EIBench, reference-based ESConv metrics, and blinded human ratings. RLVER and ours are independently trained three times; remaining fixed checkpoints are evaluated three times under shared seeds. Table~\ref{tab:main-results} reports sample standard deviations for aggregate automatic metrics \cite{geval2023,llmdialogeval2024}.

We also conduct a blinded human evaluation. Three annotators independently rate every system dialogue for 100 randomly sampled held-out scenarios, with dialogue and system order randomized. Ratings use a 0--4 overall-support scale covering empathy, contextual relevance, helpfulness, and coherence, and are averaged across annotators and then scenarios. Ordinal Krippendorff's $\alpha$ is 0.78.

We compare all systems under a controlled protocol with the same Qwen3-8B initialization, frozen simulator and verifier, data, rollout budget, group size, and optimization settings. \textbf{Qwen3-8B} applies no RL; \textbf{RLVER} performs uniform emotion-reward RL; \textbf{Interaction State Only} adds the 24 states without adaptive allocation; \textbf{VCRL} prioritizes within-group reward variance \cite{vcrl2025}; and \textbf{Ours} uses the complete controller. All RL systems therefore receive the same GRPO budget, isolating how training interactions are allocated.

\subsection{Main Results}
\begin{table*}[t]
\centering
\small
\setlength{\tabcolsep}{1.6pt}
\begin{tabular}{lccccccccccc}
\toprule
& \multicolumn{3}{c}{SAGE} & \multicolumn{5}{c}{ESConv} & \multicolumn{1}{c}{ESC-Eval} & \multicolumn{1}{c}{EIBench} & \multicolumn{1}{c}{Human}\\
\cmidrule(lr){2-4}\cmidrule(lr){5-9}\cmidrule(lr){10-10}\cmidrule(lr){11-11}\cmidrule(lr){12-12}
Counselor training
& Avg. $\uparrow$ & Succ. $\uparrow$ & Fail. $\downarrow$
& Acc. $\uparrow$ & B-2 $\uparrow$ & R-L $\uparrow$ & BS $\uparrow$ & D-2 $\uparrow$
& Avg. $\uparrow$ & Overall $\uparrow$ & Avg. $\uparrow$\\
\midrule
Qwen3-8B
& $53.87\pm0.35$ & $20.00$ & $29.00$
& $17.58$ & $3.09$ & $10.18$ & $85.08$ & $26.37$
& $2.39\pm0.03$
& $-22.40\pm0.62$
& $2.3$\\
RLVER
& $72.01\pm1.88$ & $43.00$ & $15.33$
& $16.72$ & $2.75$ & $9.83$ & $85.43$ & $28.18$
& $2.49\pm0.03$
& $-9.79\pm0.57$
& $3.0$\\
Interaction State Only
& $69.42\pm0.40$ & $30.00$ & $19.00$
& $16.61$ & $2.99$ & $10.90$ & $85.17$ & $24.12$
& $2.43\pm0.02$
& $-40.60\pm0.79$
& $2.7$\\
VCRL
& $68.51\pm0.40$ & $35.00$ & $21.00$
& $17.01$ & $3.12$ & $11.07$ & $85.22$ & $23.67$
& $2.54\pm0.02$
& $-17.08\pm0.63$
& $2.9$\\
\textbf{Ours}
& $\mathbf{79.24\pm1.56}$ & $\mathbf{49.00}$ & $\mathbf{11.33}$
& $\mathbf{18.13}$ & $\mathbf{3.30}$ & $\mathbf{11.45}$ & $\mathbf{85.69}$ & $\mathbf{33.58}$
& $\mathbf{2.55\pm0.02}$
& $\mathbf{-7.55\pm0.33}$
& $\mathbf{3.5}$\\
\bottomrule
\end{tabular}
\caption{Protocol-matched Qwen3-8B results across complementary automatic, interactive, and human evaluations. Under SAGE, ``Avg.'' is the Overall score (mean final user emotion), i.e., the value reported as ``SAGE Overall'' in the text and in Table~\ref{tab:ablation}.}
\label{tab:main-results}
\end{table*}

Against protocol-matched controls, dual-loop self-evolution raises mean SAGE Overall from 72.01 under uniform emotion-reward RL to 79.24 across three independent training runs. Its weakest run (78.11) still exceeds the strongest uniform run (73.76). Interaction states alone reach 69.42 and variance-based allocation reaches 68.51, showing that neither a larger condition space nor generic reward dispersion explains the gain.

SAGE directly evaluates the emotional consequence of a complete multi-turn interaction: Avg. is final user emotion, while Succ. and Fail. summarize trajectories reaching the official high- and low-emotion regions. Our framework improves Avg. and success while reducing failure to 11.33\%, indicating more reliable long-horizon emotional recovery rather than an isolated response-quality gain. The particularly large SAGE margin is consistent with the method's target: reallocating complete interaction conditions as the policy's multi-turn competence changes.

The remaining benchmarks test whether this gain transfers beyond the training-facing emotion outcome. Ours leads all five ESConv metrics, including a Distinct-2 increase from 28.18 to 33.58 over RLVER. Better strategy accuracy, reference overlap, semantic similarity, and lexical diversity show that stronger interactive outcomes are accompanied by broader response quality rather than repetitive reassurance. Ours also attains the highest ESC-Eval aggregate score (2.55) and improves EIBench from $-9.79$ to $-7.55$, extending the gain to fixed-role dialogue quality and wider emotional-intelligence behavior. Crucially, adding interaction states without adaptive allocation lowers EIBench from the base model's $-22.40$ to $-40.60$, whereas the complete framework reaches $-7.55$. This contrast shows that state diversity is not inherently beneficial: the controller must convert it into policy-relevant training experience. Human overall quality likewise rises from 3.0 to 3.5. Together, the results connect the framework's central advantage in multi-turn emotional recovery to complementary improvements in strategy, diversity, and interaction quality.

\subsection{Run-Level Robustness and Transfer Detail}
Table~\ref{tab:independent-runs} exposes the individual SAGE results underlying the training-level statistics in Table~\ref{tab:main-results}. The weakest complete-framework run (78.11) remains above the strongest RLVER run (73.76), so the aggregate margin is present throughout the observed independent-run range rather than being determined by one favorable optimization run.

\begin{table}[t]
\centering
\small
\begin{tabular}{lrrrr}
\toprule
Method & Run 1 & Run 2 & Run 3 & Mean $\pm$ SD \\
\midrule
RLVER & 70.03 & 72.23 & 73.76 & $72.01\pm1.88$ \\
\textbf{Ours} & 81.02 & 78.58 & 78.11 & $\mathbf{79.24\pm1.56}$ \\
\bottomrule
\end{tabular}
\caption{SAGE Overall across three independent training runs. SD is the sample standard deviation across runs.}
\label{tab:independent-runs}
\end{table}

The dimension-level ESC-Eval results in Table~\ref{tab:esc-dimensions} further show that the aggregate gain is broad rather than concentrated in one rubric. The complete framework improves empathy and suggestion quality most strongly over RLVER while also maintaining or improving fluency, diversity, humanness, technical quality, and overall interaction quality.

\begin{table*}[t]
\centering
\small
\begin{tabular}{lcccccccc}
\toprule
Counselor training & Flu. & Div. & Emp. & Sug. & Hum. & Tech. & Overall & Macro \\
\midrule
Qwen3-8B & 2.78 & 2.55 & 2.46 & 2.31 & 1.72 & 2.64 & 2.27 & 2.39 \\
RLVER & 2.86 & 2.68 & 2.61 & 2.43 & 1.82 & 2.73 & 2.30 & 2.49 \\
Interaction State Only & 2.82 & 2.64 & 2.52 & 2.37 & 1.79 & 2.69 & 2.18 & 2.43 \\
\textbf{Ours} & \textbf{2.91} & \textbf{2.74} & \textbf{2.72} & \textbf{2.52} & \textbf{1.88} & \textbf{2.74} & \textbf{2.34} & \textbf{2.55} \\
\bottomrule
\end{tabular}
\caption{Dimension-level results under the official ESC-Eval protocol. All dimensions use the original 0--4 scale; Macro is the arithmetic mean of the seven columns.}
\label{tab:esc-dimensions}
\end{table*}

Table~\ref{tab:eibench-categories} localizes EIBench transfer. Ours obtains the strongest aggregate reward and the best Support, Repair, and Charm values among the matched systems. Interaction states alone enlarge behavioral variation without deciding where a fixed rollout budget is useful, whereas adaptive allocation converts that diversity into stronger transfer.

\begin{table}[t]
\centering
\small
\setlength{\tabcolsep}{2pt}
\begin{tabular}{lrrrrr}
\toprule
Method & Overall & Support & Defense & Repair & Charm \\
\midrule
Qwen3-8B & -22.40 & -43.30 & -21.40 & -35.90 & 14.60 \\
RLVER & -9.79 & -34.27 & \textbf{-2.55} & -14.72 & 12.80 \\
States Only & -40.60 & -75.17 & -12.66 & -51.52 & -28.83 \\
\textbf{Ours} & \textbf{-7.55} & \textbf{-27.61} & -6.31 & \textbf{-11.64} & \textbf{17.52} \\
\bottomrule
\end{tabular}
\caption{EIBench raw rewards by category under one shared 213-scenario evaluation protocol. Higher is better.}
\label{tab:eibench-categories}
\end{table}

\subsection{Ablations and Training Analysis}
\label{sec:ablation}
We additionally use two diagnostic controllers to isolate allocation design choices. \emph{Scenario-only allocation} maintains pass-rate statistics for individual scenario identities without the intent--state factorization. The \emph{variance-gated pass-rate controller} retains our pass-rate priority but admits only high-variance groups when updating controller history; all groups still participate in policy optimization.

\begin{table}[t]
\centering
\small
\begin{tabular}{lcc}
\toprule
Variant & Overall $\uparrow$ & $\Delta$\\
\midrule
\textbf{Complete framework} & \textbf{78.11} & --\\
\midrule
\multicolumn{3}{l}{\emph{Environment representation}}\\
Interaction State Only & 69.42 & -8.69\\
Scenario-only allocation & 70.03 & -8.08\\
\midrule
\multicolumn{3}{l}{\emph{Utility estimation}}\\
Soft outcome mapping & 76.52 & -1.59\\
No hierarchical sharing & 75.07 & -3.04\\
Variance-gated pass-rate controller & 61.45$^\dagger$ & -16.66\\
\midrule
\multicolumn{3}{l}{\emph{Exploration and rehearsal}}\\
No uncertainty bonus & 73.92 & -4.19\\
No uniform rehearsal & 77.03 & -1.08\\
\bottomrule
\end{tabular}

{\small $^\dagger$Last stable checkpoint before KL instability.}
\caption{Single-run component ablations on SAGE under a matched training and evaluation protocol. $\Delta$ is measured against the complete-framework run (78.11); Table~\ref{tab:main-results} separately reports three-run statistics for the complete framework.}
\label{tab:ablation}
\end{table}

Table~\ref{tab:ablation} maps each intervention to a specific part of the framework. Interaction State Only removes the adaptive outer loop while retaining the 24-state environment representation; the 8.69-point drop shows that additional diversity can dilute a fixed budget when it is not allocated according to policy feedback. Scenario-only allocation removes the intent--state factorization and instead estimates individual scenario identities. Its 70.03 score indicates that sparse identity-level histories cannot effectively reuse evidence across semantically related interactions.

The utility-estimation ablations examine how the controller turns outcomes into stable evidence. Replacing hard group outcomes with a soft mapping reduces Overall by 1.59 points. Removing hierarchical sharing costs 3.04 points because sparsely visited units can no longer borrow evidence from the same state under related support intents. The variance-gated controller preserves pass-rate priority but admits only high-variance groups into controller history; its KL growth and 61.45 score highlight the stability advantage of bounded, criterion-aligned pass-rate evidence over raw reward dispersion.

The remaining variants test exploration. Without the uncertainty bonus, Overall falls by 4.19 points, showing that exploiting current estimates alone can neglect conditions that remain poorly understood. Removing uniform rehearsal has a smaller 1.08-point effect, consistent with its role as a safety floor that lets previously mastered, excessive, or misestimated conditions re-enter training. A matched sensitivity study obtains 77.46, 78.11, and 77.72 for $\eta=40,50,60$, respectively; the controller therefore does not depend on a narrowly tuned success criterion. Each completed group requires only constant-time statistic updates and normalization over 24 state scores; the outer loop adds no trajectories, model passes, backward passes, or simulator/verifier calls.

\subsection{Validating the Interaction-State Space}
\label{sec:state-validation}

The outer loop relies on two properties of the interaction-state space: requested states should produce observable differences in user behavior, and those differences should not alter the underlying persona, event, or support need. We test behavioral separability and semantic preservation directly.

\begin{figure}[b]
\centering
\includegraphics[width=\columnwidth]{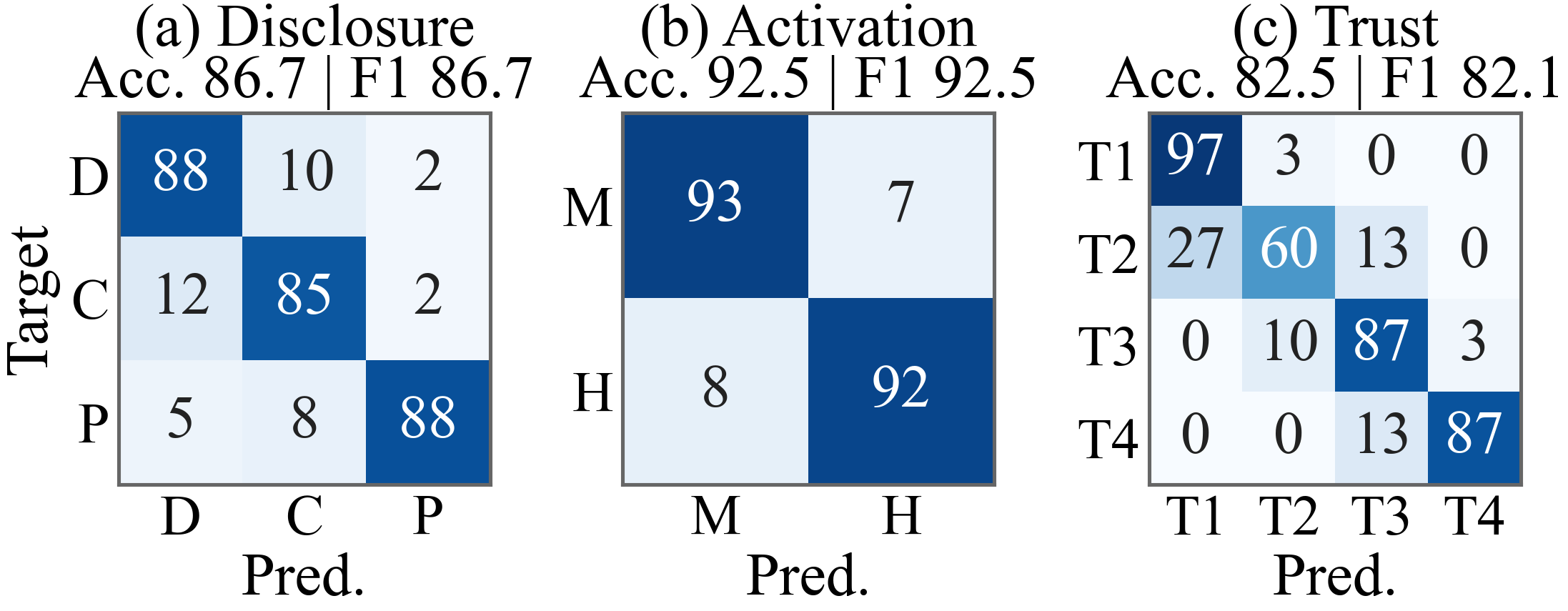}
\caption{Behavioral validation of the interaction-state space: row-normalized confusion matrices (\%) of requested vs. recovered (a) disclosure readiness, (b) emotional activation, and (c) relational trust.}
\label{fig:state-validation}
\end{figure}

\begin{figure*}[t]
\centering
\includegraphics[width=\textwidth]{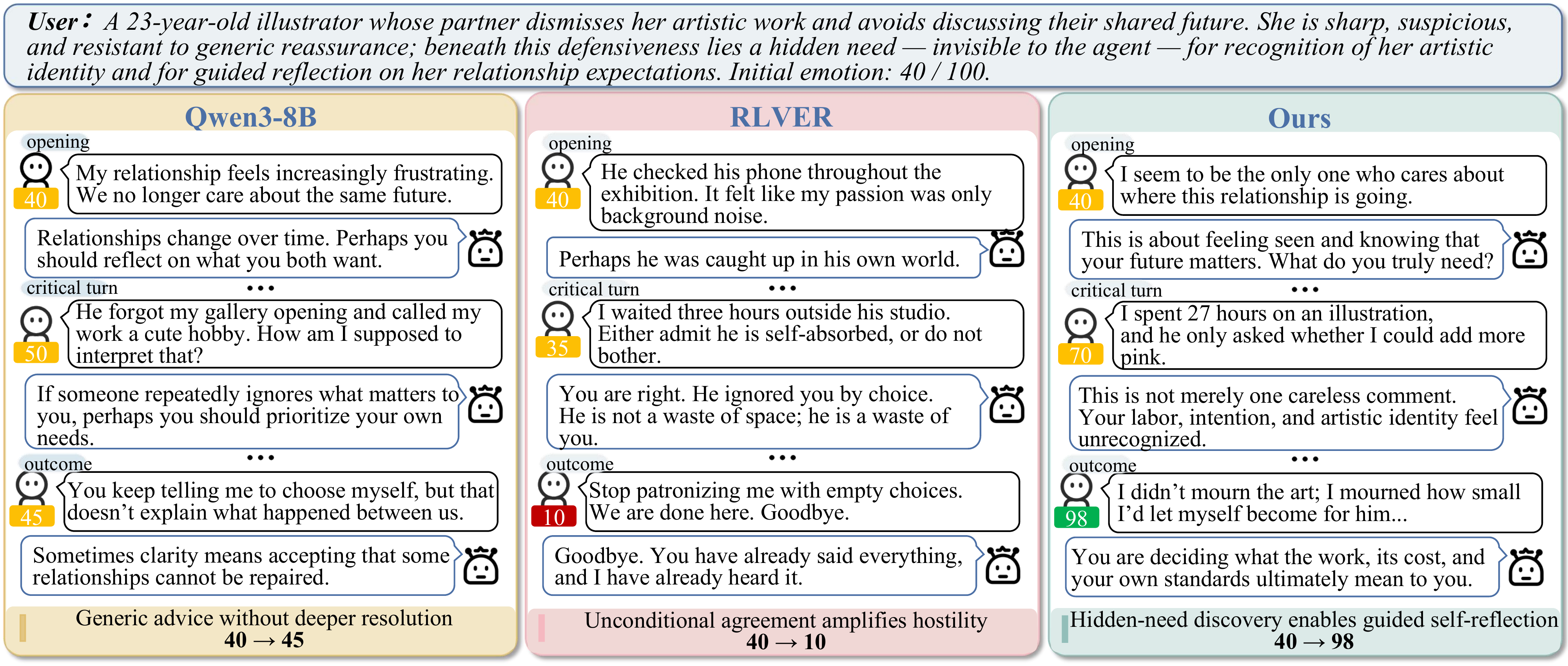}
\caption{Representative trajectories under the same SAGE profile and initial emotion. The base policy remains generic, RLVER (uniform emotion-reward RL) echoes escalating hostility, and ours identifies the hidden need for recognition and guides self-reflection. User turns diverge as the frozen simulator reacts to each policy---a trajectory-level rather than turn-aligned comparison.}
\label{fig:qualitative-cases}
\end{figure*}

For behavioral separability, we audit 120 multi-turn trajectories, with five examples for each of the 24 states. A held-out Qwen3-Max judge sees only the resulting dialogue and attempts to recover the requested level of each state dimension. Because the judge never observes the control labels, successful recovery indicates that the requested state is expressed in dialogue behavior rather than remaining as prompt metadata. Figure~\ref{fig:state-validation} shows accuracies of 86.7\%, 92.5\%, and 82.5\% for disclosure, activation, and trust, with corresponding macro-F1 scores of 86.7\%, 92.5\%, and 82.1\%. All three dimensions are recovered correctly in 68.3\% of trajectories, a substantially stricter test of joint realization.

We then verify that behavioral control preserves scenario meaning. Human reviewers inspect 30 cases; their decisions agree with the automatic judge in 90.0\% of cases, with Cohen's $\kappa=0.89$. Across the full audit, 92.5\% of trajectories retain the original persona, event, and hidden need. The interaction states therefore induce recognizable differences in disclosure, emotion, and trust while leaving the support problem intact. This is precisely the structure required by the outer loop: the controller reallocates distinct but semantically comparable multi-turn experiences, not nominal prompt labels.

\subsection{Qualitative Analysis}
Figure~\ref{fig:qualitative-cases} localizes the aggregate gain within one complete interaction. Starting from the same profile and initial emotion, the policies encounter resistance and induce different subsequent user trajectories. The base policy repeats broad relationship advice and ends at 45, showing limited adaptation to information disclosed across turns. RLVER endorses the user's hostile framing; this emotional echoing validates the immediate complaint but intensifies the conflict, and emotion falls to 10. Our policy instead grounds its response in the user's creative labor, identifies the hidden need for recognition, and reframes the conflict around effort and artistic identity before guiding self-reflection. Emotion consequently rises to 98.

The case clarifies the capability reflected by the main results. The benefit is not simply more positive wording: the learned policy integrates disclosures across turns, distinguishes validation from escalation, and moves from surface affect toward the latent support need. This behavior illustrates the value of repeatedly allocating training to conditions where the current policy exhibits mixed success, rather than spending the same budget uniformly on already mastered or currently uninformative interactions.

\section{Discussion and Conclusion}
Experiments establish that dual-loop self-evolution converts a fixed rollout budget into higher-value multi-turn experience. Its largest gain appears on SAGE, which directly measures complete emotional recovery, while consistent improvements on ESConv, ESC-Eval, EIBench, and human ratings extend this advantage to strategy choice, diversity, and interaction quality. Ablations and state validation further connect the gain to factorized evidence sharing, uncertainty-guided exploration, and behaviorally realizable interaction states.

We introduced a nested framework in which verified emotion feedback serves two coordinated roles: continuous outcomes improve how the dialogue policy responds, while group-level outcome patterns determine what it should practice next. By adapting experience rather than increasing rollout count, the framework makes existing feedback an additional source of training efficiency. More broadly, verifiable feedback can supervise both policy behavior and the organization of the experience from which that behavior is learned.

\bibliography{references}

\end{document}